\documentclass[10pt,twocolumn,a4paper]{article}

\usepackage[left=2cm, right=2cm, top=2cm, bottom=2cm]{geometry}
\usepackage{newtxtext,newtxmath} % Times Roman font (Standard for papers)
\usepackage[font=small,labelfont=bf]{caption} % Better caption formatting
\usepackage{titlesec} % Optional: Makes section titles more compact

\usepackage{authblk}

\usepackage{amsmath,amsfonts}
\usepackage{algorithmic}
\usepackage{algorithm}
\usepackage{array}
\usepackage[font=normalsize,labelfont=sf,textfont=sf]{subfig} 
\usepackage{textcomp}
\usepackage{stfloats}
\usepackage{url}
\usepackage{verbatim}
\usepackage{graphicx}
\usepackage{cite}
\usepackage{hyperref}
\hypersetup{
    colorlinks=true,
    urlcolor=blue
}

\usepackage[capitalise]{cleveref}
\usepackage[table]{xcolor}
\usepackage{booktabs}
\usepackage{multirow}
\usepackage{tabularx}
\usepackage{pifont}
\usepackage{gensymb}
\usepackage{lscape}

\newcommand{\cmark}{\ding{51}}%
\newcommand{\xmark}{{\color{lightgray}\ding{55}}}
\def\eg{\emph{e.g.}} 
\def\ie{\emph{i.e.}}

\title{RUMPL: Ray-Based Transformers for Universal Multi-View 2D to 3D Human Pose Lifting}

\author[1]{Seyed Abolfazl Ghasemzadeh}
\author[2]{Alexandre Alahi}
\author[1]{Christophe De Vleeschouwer}

\affil[1]{ICTEAM/ELEN, UCLouvain, Louvain-la-Neuve, Belgium}
\affil[2]{EPFL, Lausanne, Switzerland}

\date{} 

\begin{document}

% In 'article' class with 2 columns, \maketitle usually puts the title 
% in the left column. To span the title across the page, use this wrapper:
\twocolumn[
  \begin{@twocolumnfalse}
    \maketitle
    \begin{abstract}
      Estimating 3D human poses from 2D images remains challenging due to occlusions and projective ambiguity. Multi-view learning-based approaches mitigate these issues but often fail to generalize to real-world scenarios, as large-scale multi-view datasets with 3D ground truth are scarce and captured under constrained conditions. To overcome this limitation, recent methods rely on 2D pose estimation combined with 2D-to-3D pose lifting trained on synthetic data.
Building on our previous MPL framework, we propose \textbf{RUMPL}, a transformer-based 3D pose lifter that introduces a \textbf{3D ray-based representation} of 2D keypoints. This formulation makes the model \textbf{independent of camera calibration and the number of views}, enabling \textbf{universal deployment} across arbitrary multi-view configurations without retraining or fine-tuning. A new View Fusion Transformer leverages learned fused-ray tokens to aggregate information along rays, further improving multi-view consistency.
Extensive experiments demonstrate that RUMPL reduces MPJPE by up to \textbf{53\%} compared to triangulation and over \textbf{60\%} compared to transformer-based image-representation baselines. Results on new benchmarks, including in-the-wild multi-view and multi-person datasets, confirm its robustness and scalability. The framework's source code is available at \url{https://github.com/aghasemzadeh/OpenRUMPL}.
    \end{abstract}
    \vspace{1cm} % Add some space after abstract
  \end{@twocolumnfalse}
]

\section{Introduction}

Estimating 3D human poses, \ie~recovering the 3D positions of body joints, in a common world coordinate system is a central problem in computer vision. Its applications include sports analysis, surveillance, human tracking and re-identification, and behavioral analysis such as fall detection of the elderly and patient recovery analysis in neuroscience 
\cite{Somers_2023_WACV,bridgeman2019multi,salchow2022emerging,el2023systematic}.

Deep neural networks have helped pushing to major progress in 2D human pose estimation (HPE), but inferring 3D poses from monocular views remains fundamentally ill-posed due to projective ambiguity and occlusions. Multi-view approaches alleviate these issues by aggregating information from several cameras \cite{wang2019geometric,chen2021mvhm,srivastav2024selfpose3d,liao2024multiple:mvgformer,ma2022ppt,ma2021transfusion,wang2021mvp}. However, their training typically requires multi-view images paired with 3D ground truth-data that are scarce and limited to 'in-the-lab' settings \cite{h36m,cmu}. Consequently, most existing models fail to generalize to new scenes or camera configurations.

\textbf{Our goal} is to enable 3D human pose estimation that is ready for \textbf{'in-the-wild' deployment}, capable of handling \textbf{arbitrary scenes} and \textbf{camera setups without retraining}. This work builds upon our prior method, MPL \cite{MPL}, which addressed scene variability but remained tied to fixed camera layouts. The comparison in \cref{table:comparison-sota} summarizes the main properties of different methods, highlighting the step forward achieved by RUMPL in terms of universality.

\begin{table}[t!]
\centering
\caption{Comparison of methods and their key properties.}
\resizebox{\columnwidth}{!}{%
\renewcommand{\arraystretch}{1.2}
\begin{tabular}{l|c|c|c|c}
\toprule
\multirow{2}{*}{Method} & Recovery of & \multirow{2}{*}{New scenes} & Compatibility to  & Good accuracy  \\ 
                        & occluded keypoints & & new camera setup  & with 2 views \\ \midrule
Triangulation (baseline) & $\times$ & $\checkmark$ & $\checkmark$ & $\times$ \\ 
Learn. tri. \cite{iskakov2019learnable} & $\checkmark$ & $\times$ & $\times$ & $\times$ \\ 
PPT \cite{ma2022ppt} & $\checkmark$ & $\times$ & $\checkmark$ & $\checkmark$ \\
Gen. tri. \cite{bartol2022generalizable} & $\checkmark$ & $\times$ & $\checkmark$ & $\times$ \\
MVGFormer \cite{liao2024multiple:mvgformer} & $\checkmark$ & $\times$ & $\checkmark$ & $\times$ \\ \hline
MPL (our previous) \cite{MPL} & $\checkmark$ & $\checkmark$ & $\times$ & $\checkmark$ \\ \midrule
\textbf{RUMPL (ours)} & \textbf{$\checkmark$} & \textbf{$\checkmark$} & \textbf{$\checkmark$} & \textbf{$\checkmark$} \\ 
\bottomrule
\end{tabular}%
\label{table:comparison-sota}
}
\end{table}

Following the 2D-to-3D lifting paradigm \cite{MPL,martinez2017simple,kadkhodamohammadi2021generalizable}, we first extract 2D keypoints independently from all available images using an off-the-shelf robust 2D pose estimator, and then lift them to 3D via a dedicated network. While triangulation can serve as a simple baseline, it fails under occlusions or missing views. Previous learned fusion methods, including MPL, rely on image or 2D pose features, which depend on camera calibration or the number of views.

To overcome these limitations, we propose the \textbf{Ray-based Universal Multi-view Pose Lifter (RUMPL)}, a transformer-based model that lifts multi-view 2D poses into 3D world coordinates using a \textbf{3D ray representation}. Each 2D keypoint is represented as a ray in world space, enabling explicit geometric reasoning and removing the need to learn projection relationships. A new \textbf{View Fusion Transformer} aggregates information along rays via a learned fused-ray token, allowing RUMPL to operate with an arbitrary number of input views.

Training relies on synthetic 2D–3D pairs generated from random camera configurations using the AMASS dataset \cite{AMASS:ICCV:2019}. This enables learning from diverse, randomized setups and achieves true universality across acquisition conditions.

\textbf{Our main contributions are:}
\begin{itemize}
    \item[(i)] We propose \textbf{RUMPL}, a transformer-based 3D HPE model that generalizes to arbitrary scenes and camera setups \textbf{without retraining or fine-tuning}.
    \item[(ii)] We introduce a \textbf{3D ray representation} that encodes geometric relations explicitly, eliminating the need to learn camera calibration parameters.
    \item[(iii)] We train on \textbf{synthetic 2D–3D pose pairs} generated under random viewpoints, enabling scalable, camera-agnostic learning for real-world deployment.
\end{itemize}

\section{Related Work}
\label{sec:related}
\subsection{Monocular 3D Pose Estimation}

The current state-of-the-art in single-view 3D HPE can be categorized into two main categories. 
The first category consists of methods that infer the 3D coordinates directly from images using convolutional neural networks. These methods are complex and, more importantly, limited to the in-the-lab scenes \cite{sun2018integral,pavlakos17volumetric}.

The second category includes methods that use high quality 2D pose estimators and turn the 2D poses into 3D poses with deep neural networks, \eg, convolutional, recurrent, and transformers. These methods mostly rely on temporal consistency. 
They are simple, fast, and can be trained on motion capture datasets making them more accurate in real-world scenes compared to the works from the first category \cite{motionbert2022,Shan_2023_ICCV_d3dp,Zhang_2022_CVPR_mix_ste,pose_former_v2:2023,Kanazawa_2018_CVPR}. However, due to working with a single view, they are unable to predict the 3D poses in world coordinates.

\subsection{Multi-view 3D Pose Estimation}
In recent years, there has been an increasing interest in dealing with multi-view 3D HPE by fusing the image features from all views \cite{ma2022ppt,ma2021transfusion,epipolartransformers,multiviewpose_cross_view_fusion,iskakov2019learnable}. 
Epipolar Transformers \cite{epipolartransformers} fuses the features of one pixel in one view with features along the corresponding epipolar line of the other views. 
Cross View Fusion \cite{multiviewpose_cross_view_fusion} learns a fixed attention matrix for fusing features in all other views. 
TransFusion \cite{ma2021transfusion} applies transformers globally to fuse features of the reference views. 
Learnable-triangulation \cite{iskakov2019learnable}  aggregates the features from different views in a volume by unprojecting the features using the camera calibration parameters. 
Finally, PPT \cite{ma2022ppt} utilizes a transformer-based approach for 2D pose estimation and makes use of the attention matrix weights to prune the unimportant visual tokens.
However, all these methods require multi-view images paired with annotated 3D pose for training, making them suited to images corresponding to the scene and acquisition setup used at training but vulnerable to novel deployment conditions.

To alleviate this limitation, the authors in \cite{kadkhodamohammadi2021generalizable} propose concatenating all the joints' 2D coordinates from all views and treating them as a single input to a fully connected network trained to predict the 3D pose in world coordinates. However, the results show (see \cref{table:mpjpe-h36m}) that a fully connected network is prone to strong over-fitting and can fall short in accuracy when facing unseen data. 
SelfPose3d \cite{srivastav2024selfpose3d} proposes a self-supervised approach which does not rely on any 2D or 3D ground-truth pose. However, it needs to be retrained every time the scene or its camera setup changes.
Generalizable Triangulation \cite{bartol2022generalizable} introduces a stochastic framework for human pose triangulation.
MVGFormer \cite{liao2024multiple:mvgformer} proposes a hybrid model which has a series of geometric and appearance modules organized in an iterative manner.
Even though they generalize well to new scenes and camera changes, they require higher number of views to give good accuracy and do not work well with a low number of views.

In our previous work, MPL \cite{MPL}, we considered a transformer to fuse 2D keypoints from multiple views. However, MPL is limited to a specific camera setup, and it requires retraining each time the camera locations change. 
In contrast, thanks to our novel Ray-based representation of keypoints, our RUMPL method succeeds in training a model that is agnostic to the scene and camera acquisition setup.
\cref{table:comparison-sota} compares the various properties of different methods explained here.

\subsection{3D Ray Representation}

Ray3D \cite{zhan2022ray3d} introduces a ray representation of keypoints to remove the dependency on camera intrinsic parameters. 
Our work extends Ray3D in two important aspects: (i) our ray represents a 2D image point in a unique (common to all cameras) 3D world referential, while Ray3D did consider the ray to circumvent  intrinsic parameters (and thus defines the ray in each camera referential), while feeding the network with explicit extrinsic parameters; (ii) our work considers a multi-camera setup, and our motivation to adopt a world-coordinates ray representation is to train a model that remains valid whatever the intrinsic and extrinsic camera parameters.

\subsection{Synthetic Pose Generation}
Generating synthetic 2D-3D pairs has been a line work for a long time. Works in \cite{li2020cascaded}, PoseAug \cite{gong2021poseaug}, and AdaptPose \cite{gholami2022adaptpose} focus on interpolating new poses in the 3D pose manifold while the contribution of our pose generator is to use 3D mesh to create 2D pose from rendered meshes.

\section{RUMPL: Ray-based Universal Multi-view Pose Lifter}
\label{sec:methodology}

% \subsection{Universal Multi-view 3D Pose Lifter}
% \label{ssec:mpl}

% \begin{figure*}[!t]
% \centering
% \includegraphics[width=\textwidth,trim={0.5cm 8.5cm 0.5cm 9cm},clip]{images/network.pdf}
% \caption{RUMPL takes 3D rays of all keypoints from all views, fuses the rays associated to a given keypoint and then jointly considers the whole set of keypoints using a transformer network.}
% \label{fig:network}
% \end{figure*}

\begin{figure*}[!t]
\centering
\includegraphics[width=\textwidth,trim={0.5cm 6.5cm 0.3cm 6.5cm},clip]{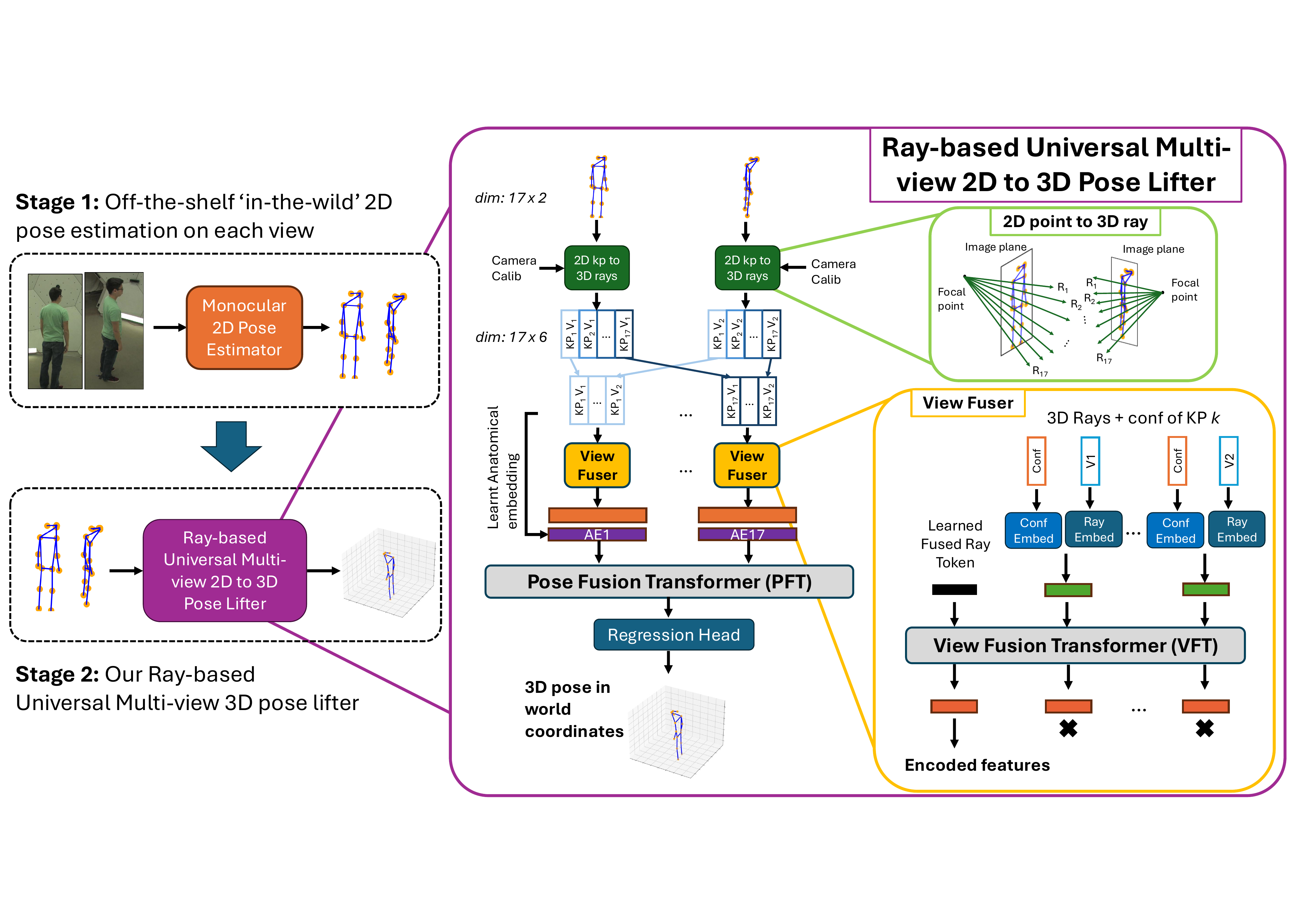}
\caption{RUMPL takes 3D rays of all keypoints from all views, fuses the rays associated to a given keypoint and then jointly considers the whole set of keypoints using a transformer network.}
\label{fig:network}
\end{figure*}

% \begin{figure*}
% \centering
% \includegraphics[width=\textwidth,trim={14cm 11cm 11cm 5cm},clip]{images/network_vertical.pdf}
% \caption{RUMPL takes 3D rays of all keypoints from all views, fuses the rays associated to a given keypoint and then jointly considers the whole set of keypoints using a transformer network.}
% \label{fig:network}
% \end{figure*}

As shown in \cref{fig:network}, our 3D human pose estimation pipeline is composed of two stages. First, the images from $N$ different views are fed to an off-the-shelf monocular 2D pose estimator to get $N$ skeletons made of $M$ 2D keypoints.
Second, the 2D poses are transformed to 3D rays using the provided camera calibration, and fed to the 2D-to-3D pose lifter.
% to get a 3D pose.
% Please see the supplementary material for a detailed schematic of RUMPL.

\subsection{2D Pose Keypoints}
RUMPL takes as input the 2D pose skeletons predicted by an off-the-shelf 2D pose estimator for $N$ views. Let $P_{i} \in \mathbb{R}^{M \times 2}$ and $C_{i} \in \mathbb{R}^{M}$ denote the $M$ pairs of coordinates defining the location and the confidence (as predicted by the off-the-shelf 2D pose estimator) of the $M$ keypoints in the $i^{th}$ view, respectively. RUMPL takes $\{ P_{i} \}_{i<N}$ as input and predicts the 3D pose skeleton $Q \in \mathbb{R}^{M \times 3}$ in the world coordinate system. RUMPL includes three main components, namely the Keypoint to Ray Converter, the View Fusion Transformer (\emph{VFT}), and the Pose Fusion Transformer (\emph{PFT}), defined as follows.

\subsection{Keypoint to Ray Converter}
\label{ssec:ray}

Training a network compatible with new camera setups is not a trivial task. 
% We show in \cref{importance-ray} that feeding the network directly with 2D keypoints (in pixel space) is not sufficient for training a network compatible to arbitrary camera setups.
We were not successful feeding the network with 2D keypoints along with camera calibration (intrinsic and extrinsic) information (see \cref{table:importance-ray}).
% However, it seems that formulating the problem this way does not yield good results. 

Therefore, we propose to adopt 3D rays to represent 2D keypoints, thereby alleviating the need for the network to the relation between 2D coordinates and camera calibration.
This ray representation consists of a vector indicating the direction of the ray and a point, \ie, the camera location, on the ray. 
Let $P \in \mathbb{R}^{M \times 2}$ denote the $M$ pairs of coordinates defining the location of the $M$ keypoints in the pixel space. 
Intrinsic parameters of the camera are defined by $K$ ($3 \times 3$). Extrinsic parameters of the camera are defined by $R$ ($3\times3$) and $T$ ($3\times1$) containing the rotation matrix and the location of camera, respectively.
For converting the 2D pixel coordinates to 3D rays, we follow the below steps.
\begin{enumerate}
    \item First, we convert the 2D pixel points into homogeneous coordinates by appending 1:

    \begin{equation}
  P_{hom} = [P~~1].
  \label{eq:hom}
\end{equation}

This results in an $M \times 3$ matrix, that is to transform to normalized camera coordinates by applying the inverse of the intrisic matrix:

\begin{equation}
    P_{norm} = K^{-1}P^T_{hom}.
\end{equation}
This gives a $3 \times M$ matrix where each column represents a point in normalized camera coordinates.

    \item We compute, in the world referential, the direction of each ray originated from the camera center $T$. 

\begin{equation}
    D = R^TP_{norm}.
\end{equation}
This gives a $3 \times M$ matrix where each column represents a direction corresponding to a keypoint.

\item Finally, we concatenate the directions and the camera locations to represent the 3D ray for each keypoint.
\begin{equation}
    \mathcal{R} = [T~|~D^T].
\end{equation}
\end{enumerate}

This module takes as input the keypoint coordinates $P_{i} \in \mathbb{R}^{M \times 2}$ and the camera calibration matrices $K$, $R$, and $T$ and returns 3D rays $\mathcal{R}_{i} \in \mathbb{R}^{M \times 6}$.

\subsection{Keypoint-level View Fusion Transformer (VFT)}

VFT aggregates the information associated to a given keypoint (from all views) independently of other keypoints. It is identical for all keypoints and encodes the keypoints in a format suited to the subsequent 3D pose estimation module.
Let $\mathcal{R}_{j} \in \mathbb{R}^{N \times 6}$ denote the $N$ 3D encoded rays from all views for the $j^{th}$ keypoint. 

VFT takes $\mathcal{R}_{j} \in \mathbb{R}^{N \times 6}$ and $\mathcal{C}_{j} \in \mathbb{R}^{N}$ as input and returns a $D$-dimensional hidden embedding for each view.
This is done as follows; first, a linear projection is applied to project each encoded ray to a $D/2$-dimensional vector. Another linear projection is applied to project confidence $C_j$ to a $D/2$-dimensional vector. These two vectors are then concatenated to shape a $D$-dimensional vector, resulting into $X_{j}(i) \in \mathbb{R}^{D}$ for the $i^{th}$ view.

Inspired by the usage of the transformers in classification tasks \cite{dosovitskiy2020image}, we consider a learnable ray token $X_j(0) \in \mathbb{R}^{D}$ called fusion token which is expected to contain the fused features at the output of the transformer.
It is worth noting that this approach gives RUMPL the ability to accept any number of views without the need for retraining. This is because the architecture, in no way, is dependent of the number of views.

Next, the $N$ ray tokens associated to a keypoint are fed to an encoder transformer that returns $N$ vectors $X_{j}^V(i) \in \mathbb{R}^{D}$, with $0 \leq i \leq N$, for keypoint $j$.
We adopt the same architecture as in \cite{pose_former:2021} for the encoder transformer layers.
It relies on multi-headed self attention (MHSA) and multi-layer perceptron (MLP). The transformer network has $L$ layers and each layer has $H$ heads \cite{attn_all_need}.

Finally, the fusion token $X_{j}^V(0)$ for $j^{th}$ keypoint is picked as the fused features of the corresponding keypoint resulting in $Y(j) \in \mathbb{R}^D$ for keypoint $j$.

\subsection{Pose Fusion Transformer (PFT)}

PFT takes the tokens fused by VFT as $Y(j) \in \mathbb{R}^D$ for each keypoint $j$, with $1 \leq j \leq M$. An anatomical embedding $AE(j) \in \mathbb{R}^{D}$, learned for each keypoint $j$, is first added to each $Y(j)$.
This is expected to make the transformer aware of the the joints type information \cite{attn_all_need}, resulting in a keypoint token $Y^\prime(j)=Y(j)+AE(j)$ for the $j^{th}$ keypoint.
Next, the keypoint tokens are fed to an encoder transformer, which returns a fused embedding $Y^P(j) \in \mathbb{R}^{D}$.
The transformer encoder in PFT follows the same scheme as that of VFT. Eventually, $Y^P$ is passed to a regression head to make the final output, \ie, the 3D pose skeleton $Q \in \mathbb{R}^{M \times 3}$ in the world coordinate system.
The regression head is composed of a 1-layer MLP.

\subsection{Loss Function}

To train RUMPL, the output is directly compared to the ground truth 3D pose using Mean Per Joint Position Error (MPJPE).
Eq. \eqref{eq:mpjpe} shows how MPJPE is calculated, with $Q$ and $\hat{Q}$ denoting the predicted and the ground truth 3D pose, respectively.

\begin{equation}
  \text{MPJPE} = \dfrac{1}{M} \sum_{j=1}^{M} \|{Q_{j} - \hat{Q_{j}}} \|_2.
  \label{eq:mpjpe}
\end{equation}

\section{MHP: Mesh-based 2D-3D Human Pose Dataset Generator}
\label{sec:method_dataset_pipeline}

\begin{figure*}[!t]
\centering
\includegraphics[width=\textwidth,trim={0cm 14.25cm 0cm 18.5cm},clip]{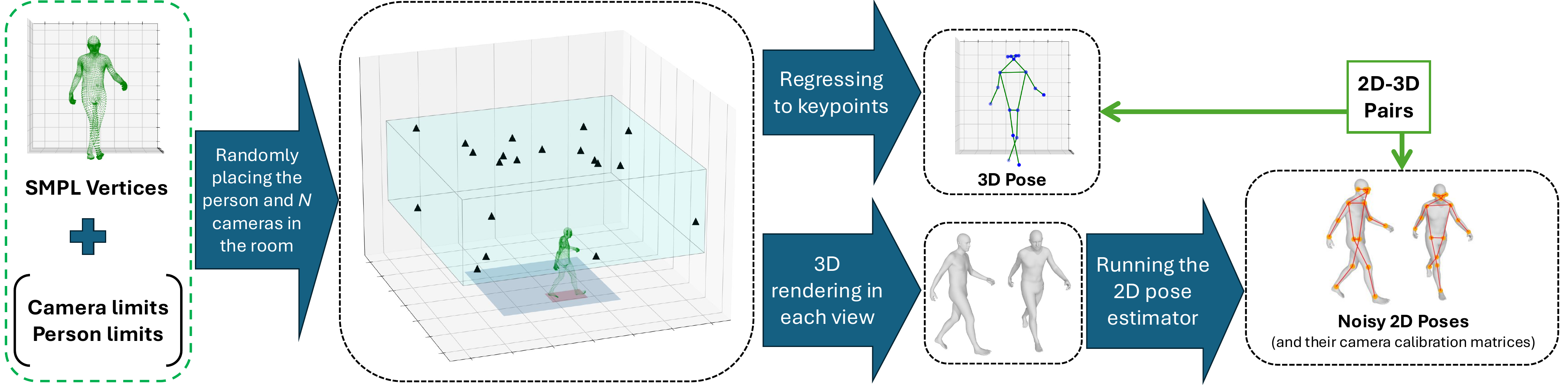}
\caption{Our \textbf{Mesh-based Human Pose Dataset Generator (MHP)} takes 3D mesh vertices and the limits of camera and person's displacement in a given room as inputs. It randomly positions the mesh and $N$ cameras in the scene and returns the 3D pose ground truth and one 2D pose per camera view. Each 2D pose is estimated by an off-the-shelf 2D pose estimation model, applied to the image of the mesh rendered in the corresponding view. The 3D keypoint regressor defines the 3D pose ground truth in a way that is consistent with the definition of ground truth at testing. This makes the 2D-3D pairs of human pose appropriate for training an accurate and robust RUMPL, able to turn the 2D poses computed by the off-the-shelf pose estimator into a 3D skeleton that is consistent with the 3D keypoints expected at inference. Note that the triangles represent the location of the cameras, the light blue cube illustrates the camera limits, and the blue surface depicts the person's limits in the room.}
\label{fig:mhp}
\end{figure*}

The most natural way to generate the 2D-3D pairs required to train our RUMPL is to randomly position a 3D skeleton in arbitrary positions within the observed 3D space and a number of cameras in the same scene, project this 3D skeleton into a view of interest, thereby generating the corresponding 2D skeleton.  
This strategy, however, suffers from two important drawbacks. First, it appears that the 3D skeleton datasets adopt distinct rules to define the keypoints. Most often, those definitions are not compatible with the ones adopted by the off-the-shelf 2D pose estimation model used to process 2D images at inference, making the 2D-3D associations considered at training incompatible with the 2D pose manipulated at inference. 
Second, this natural 3D skeleton projection strategy is unable to mimic the noise and errors affecting 2D pose estimation at inference.

To alleviate those issues, following \cite{MPL} and \cite{maiorca2024selfavataranimationvirtualreality}, instead of considering 3D skeleton and their 2D projections to train our RUMPL, we propose to generate the 2D-3D pairs of skeletons from meshes.
Therefore, we make use of AMASS \cite{AMASS:ICCV:2019}, an archive providing a variety of human shapes collected from many motion capture datasets and represented with standard 3D mesh models, convertible to SMPL vertices \cite{SMPL:2015} that can be transferred to keypoints by regression \cite{li2021hybrik}.

Considering a dataset of 3D meshes offers two valuable properties. First, the 2D pose associated with a 3D mesh can be created by applying the off-the-shelf 2D pose estimation model used at inference on synthetic 2D renderings of the 3D mesh. This makes the 2D-to-3D pose lifter robust to noise induced by the 2D pose estimator. Second, the regressor used to convert the 3D mesh into a 3D skeleton can be designed to be compatible with the definition of keypoints adopted by the 2D pose estimator.

\cref{fig:mhp} shows the pipeline of our proposed Mesh-based Human Pose Dataset Generator (MHP) for 2D-3D pairs of human skeletons generation. It depicts how 
the 3D poses are generated by regressing each AMASS 3D mesh to an $M$-joint skeleton.
Given the calibration parameter of a camera, the 3D mesh is rendered in the image space
using the open-source softwares \textbf{Body Visualizer}~\cite{SMPL-X:2019} and \textbf{Trimesh} \cite{trimesh}. Then the rendered images are passed through the 2D pose estimator, \ie, the same as the one used at inference, to obtain the $M$-joint 2D poses. 
Finally, the predicted 2D poses from different views are paired with the regressed 3D keypoints to be used during training. An overview of this process is shown in \cref{fig:mhp}.

It is worth mentioning that the camera locations are picked randomly within a pre-defined range for every 3D pose to increase the diversity of the projective observations.
Thanks to utilizing the 3D ray representation of the 2D keypoints in RUMPL, the keypoints' representations are independent of the camera calibration resulting in a universal trained model, \ie, valid for arbitrary camera parameters (within the range used at training). 
This is in contrast to MPL \cite{MPL} requiring to be trained specifically for each acquisition setup.

\section{Experimental Validation}
\label{sec:exp}

\subsection{Evaluation Metric and Datasets}
\label{subsec:eval}

The 3D pose is evaluated by MPJPE (see Eq. \eqref{eq:mpjpe}), computed in world coordinates in \textit{mm}, between the ground truth 3D pose and the predicted 3D pose. 
We evaluate our RUMPL and alternative methods on three commonly used 3D human pose datasets, Human3.6M \cite{h36m}, CMU Panoptic \cite{cmu}, and RICH \cite{rich}. In all our experiments, no image or 3D pose of the evaluation dataset is used to design or train the tested method.

\subsubsection{Human3.6M} 
It is the most widely used public dataset for single-person 3D HPE. It consists of 11 professional actors performing 17 actions such as talking, walking, and sitting. The videos are recorded from 4 different views in an indoor environment. In total, this dataset contains 3.6 million video frames with 3D ground truth annotations, collected by an accurate marker-based motion capture system. Prior works \cite{li2021hybrik,ma2022ppt,pose_former:2021}, utilized all the 15 actions for training their models on five subjects (S1, S5, S6, S7, and S8) and tested them on two subjects (S9 and S11). We consider the same two subjects for testing our network, but do not rely on Human3.6M images or 3D poses to train our RUMPL model.
For evaluating our RUMPL on Human3.6M, we generate the RUMPL training 2D-3D pose dataset using our mesh-based human pose (MHP) generator.
% (see \cref{sec:method_dataset_pipeline}).

It is worth noting that the definition of keypoints in the Human3.6M dataset differs from the one introduced by COCO \cite{coco:Lin2014MicrosoftCC} and adopted by the off-the-shelf 2D pose estimator used in the study. 
Following \cite{MPL}, we have improved the consistency between the two formats by averaging the location of several COCO keypoints. For instance, averaging the \textit{ears} and \textit{eyes} locations from the COCO format results in a keypoint that closely matches the \textit{head} keypoint in the Human3.6M format. Similarly, averaging the \textit{hips} in COCO provides the \textit{pelvis} in Human3.6M, and the mean of the \textit{neck} and the newly created \textit{pelvis} results in a \textit{torso} keypoint.

\subsubsection{CMU Panoptic} 
This dataset is a multi-view public dataset made available by the Carnegie Mellon University. The dataset contains Full-HD video streams of 40 subjects captured/observed from up to 31 cameras located in a dome. 3D ground-truth pose annotations are generated via triangulation using all camera views. Following previous works \cite{iskakov2019learnable,Xiang2018MonocularTC}, we use the 17-keypoint subset (from the 19-joint annotation format) that are in line with the popular COCO format. 
For evaluating on CMU, we generate the RUMPL training 3D-2D pose dataset utilizing our MHP with AMASS meshes. Note that the CMU motion capture that is part of the AMASS collection has not been included in any of our training datasets. Finally, we test our RUMPL on the test split recommended in \cite{Xiang2018MonocularTC}, where only single-person scenes are selected.

\subsubsection{RICH} 
Real scenes, Interaction, Contact, and Humans dataset \cite{rich} is a dataset of multi-view videos with accurate bodies and 3D scans using marker-less motion capture systems. We extract 3D ground truth keypoints from the provided SMPL-x meshes for our evaluation.
For evaluating on RICH, we generate the RUMPL training 3D-2D pose dataset utilizing our MHP with AMASS meshes.

\subsection{Implementation Details}
\label{sec:implementation}

For predicting the 2D pose in the image space, we use the open-source pose estimator framework \textbf{MMPOSE} \cite{mmpose2020}. 
It involves the popular HRNet \cite{sun2019deep} architecture, pretrained on COCO
\cite{coco:Lin2014MicrosoftCC}.
The average precision (AP) \cite{coco:Lin2014MicrosoftCC} of the MMPOSE 2D output on AMASS rendered images 80.0\%, 93.0\%, and 84.7\% when applied with person and camera displacements corresponding to the ones observed in the Human3.6M, CMU, and RICH datasets, respectively.
Note that AP is computed when Object Keypoint Similarity (OKS) \cite{coco:Lin2014MicrosoftCC} threshold is set to 0.75 \footnote{Please refer to \cite{coco:Lin2014MicrosoftCC} for more information about OKS and AP.}. This shows that the chosen off-the-shelf 2D pose estimator 
is able to effectively handle 3D rendered human mesh images.

We set $M$, \ie, the number of joints, to 17. The hidden embeddings dimension $D$ is set to 256 in our RUMPL. Both VFT and PFT have the same configuration parameters. 
The number of encoder heads $H$ is set to 8, and the number of encoder layers $L$ is set to 12 for all the experiments.
During RUMPL training, the input batch size is set to 32, and the network is optimized using the Adam optimizer \cite{Kingma2014AdamAM} with MPJPE loss, for 20 epochs. The learning rate is initialized at 0.0001, and decays at the 10th and 15th epochs with a decreasing factor of 0.1.

\subsection{Comparison to Multi-view Prior Works}
\label{ssec:results}

Since our contribution primarily aims to enable 3D pose estimation in arbitrary observation conditions, a fair positioning should compare our RUMPL to works that are either not trained (typically because they triangulate the 2D pose estimation) or trained on other datasets than the one used at testing. 
Alternatives with publicly available code and trained weights are rare among previous works. We have identified four methods, namely Learnable-triangulation \cite{iskakov2019learnable}, TransFusion \cite{ma2021transfusion}, PPT \cite{ma2022ppt}, MVGFormer \cite{liao2024multiple:mvgformer}, and MPL \cite{MPL}.

Learnable-triangulation \cite{iskakov2019learnable} aggregates features extracted from the image spaces by unprojecting them into a common voxel space. The publicly available data only provides the trained weights on Human3.6M; thus we test their network on CMU and RICH dataset.
TransFusion \cite{ma2021transfusion} trains a transformer-based network that, by globally attending features in the image space, fuses the information from different views.
As the trained weights are only provided on Human3.6M, we tested their network on CMU and RICH where it completely failed to output any meaningful prediction. We believe this is caused by the global attention mechanism making it dependent of the scenery features. 
Next, PPT \cite{ma2022ppt} follows the same approach as TransFusion but, instead of considering a global attention process in the fusion process, it prunes the unimportant tokens. Trained weights are only provided on Human3.6M. 
However, since PPT is more recent and provides more accurate results than TransFusion \cite{ma2022ppt}, we follow their guidelines to train their network on CMU dataset and also test on Human3.6M. It is worth mentioning that we also tested their network on RICH dataset when trained on Human3.6M where it failed to output any meaningful pose.
We also compare RUMPL to MVGFormer \cite{liao2024multiple:mvgformer}. As the authors provide weights trained on CMU dataset, we tested MVGFormer on Human3.6M.

Finally, we compare our RUMPL to MPL \cite{MPL} in two different variants. First, we keep the MPL architecture, feed it with 2D keypoints in pixel space, and second, we change the input tokens to our proposed 3D rays (same way as in RUMPL) and feed it to the MPL architecture.
The reason behind this comparison is to evaluate the impact of the changes affecting the RUMPL architecture compared to the MPL one, beyond the input format, and to show the importance of the proposed architecture in the accuracy of the 3D pose estimation.
We also compare our RUMPL to a 3D pose lifter implemented based on a Fully Connected network, following the guidelines in \cite{kadkhodamohammadi2021generalizable} (with 3D ray formatting of the inputs considered for RUMPL), instead of our transformer-based solution.

\subsubsection{Human3.6M}
% \textbf{Human3.6M}:
\cref{table:mpjpe-h36m} shows the 3D HPE results on the test set of Human3.6M for each action. 
To evaluate the impact of the mismatch between the COCO and Human3.6M formats
% (see \cref{subsec:eval})
on the evaluation metrics, the last column in \cref{table:mpjpe-h36m} computes the MPJPE only on the keypoints whose definitions are consistent in both the 3D and 2D formats. This set of keypoints is denoted KP* and corresponds to knees, ankles, shoulders, elbows, and wrists.

The results show that PPT \cite{ma2022ppt} is unable to generalize well to unseen images. Despite a higher number of views, it falls short both compared to the triangulation baselines and to our RUMPL. 
Moreover, MVGFormer \cite{liao2024multiple:mvgformer} stays par with triangulation when utilizing 2 views, and even when enjoying 4 views, it falls short compared to our RUMPL.
This explains why triangulation is the reference when 3D pose estimation has to be deployed 'in-the-wild'. 

The results show that utilizing MPL and image coordinate tokens as inputs fails compared to other methods. However when using the MPL architecture coupled with the 3D ray tokens as input, we see an improvement of 37.6 $mm$ in the MPJPE (KP*) on Human3.6M val set compared to triangulation.
This improvement becomes even more when utilizing RUMPL architecture where it improves the MPJPE (KP*) by 57.9 $mm$ on average when using two views. This corresponds to an MPJPE reduction of about $50.5\%$. This gain raises to $56.7\%$ when accounting for keypoints for which the 3D ground truth definition is not consistent with the one adopted by the 2D pose estimator. This reveals the ability of our RUMPL to predict 3D keypoints that (moderately) differ from the keypoints predicted by the off-the-shelf 2D pose estimator, thereby increasing the flexibility of our framework\footnote{For example, 3D keypoints could be defined as internal body points, even if the associated 2D keypoints are defined based on visual, but 3D-inconsistent, cues in the projected views.}.

Moreover, \cref{table:mpjpe-h36m} shows the importance of the utilization of keypoint confidence (from off-the-shelf 2D pose estimator) in our architecture. The results show that not using the confidence results in an increase of 8.1 $mm$ in MPJPE (KP*).
Finally, \cref{table:mpjpe-h36m} shows that replacing the RUMPL transformer by a fully connected network (leading to a solution close to \cite{kadkhodamohammadi2021generalizable}) results in a significantly worse pose lifting.

%%%%%%%%%%% Table with only 2 cameras
\begin{table*}[h]
\centering
\caption{3D MPJPE in $mm$ on the Human3.6M test set for different methods. The last but one column averages the MPJPE on the multiple actions considered in the test set. The last column does the same but only considers the subset of keypoints KP* for which the 3D ground truth definition (adopted by Human3.6M) is consistent with the definition adopted by the MMPOSE 2D pose estimator. Bold numbers highlight the best MPJPE results for two-view setups. RCS refers to training with randomly sampled camera configurations using MHP. RUMPL always uses RCS. 'w/o Conf' means RUMPL without 2D pose confidence.}
\label{table:mpjpe-h36m}
\resizebox{\textwidth}{!}{
\begin{tabular}{l|c|*{15}{c}|c c}  % Using the new column type P with 1.2cm width
\toprule
\multirow{3}{*}{Method} & \multirow{3}{*}{\begin{tabular}[c]{@{}c@{}}\#\\ Views\end{tabular}} & \multirow{3}{*}{Dir.} & \multirow{3}{*}{Dis.} & \multirow{3}{*}{Eat.} & \multirow{3}{*}{Gre.} & \multirow{3}{*}{Phn.} & \multirow{3}{*}{Pht.} & \multirow{3}{*}{Pose} & \multirow{3}{*}{Pch.} & \multirow{3}{*}{Sit.} & \multirow{3}{*}{\begin{tabular}[c]{@{}c@{}}Sit.\\ Down\end{tabular}} & \multirow{3}{*}{Smo.} & \multirow{3}{*}{Wait} & \multirow{3}{*}{\begin{tabular}[c]{@{}c@{}}Walk\\ Dog\end{tabular}} & \multirow{3}{*}{Walk} & \multirow{3}{*}{\begin{tabular}[c]{@{}c@{}}Walk\\ Two\end{tabular}} & \multicolumn{2}{c}{Avg.} \\ 
\cmidrule{18-19}
& & & & & & & & & & & & & & & & & (All KP) & (KP*) \\
\midrule
PPT \cite{ma2022ppt} & 4 & 677.5 & 162.8 & 115.0 & 209.3 & 136.7 & 286.6 & 177.4 & 226.9 & 253.3 & 218.6 & 179.2 & 99.7 & 81.2 & 155.6 & 98.3 & 196.2 & 274.41 \\
\multirow{2}{*}{MVGFormer~\cite{liao2024multiple:mvgformer}} & 2 & - & - & - & - & - & - & - & - & - & - & - & - & - & - & - & - & 124.7 \\
& 4 & - & - & - & - & - & - & - & - & - & - & - & - & - & - & - & - & 67.3 \\
\midrule
\multirow{1}{*}{Triangulation} & 2 & 157.4 & 121.4 & 81.8 & 149.3 & 130.3 & 105.5 & 101.7 & 201.6 & 98.4 & 114.4 & 107.5 & 175.8 & 73.1 & 88.6 & 80.0 & 121.2 & 114.7 \\
\midrule
% \midrule
\multirow{2}{*}{\begin{tabular}[t]{@{}l@{}}
     FullyConnected~\cite{kadkhodamohammadi2021generalizable}  \\
     + RCS + Rays 
\end{tabular}} & \multirow{2}{*}{2} & \multirow{2}{*}{111.7} & \multirow{2}{*}{104.9} &  \multirow{2}{*}{106.4} &  \multirow{2}{*}{129.2} &  \multirow{2}{*}{142.3} &  \multirow{2}{*}{91.5} &  \multirow{2}{*}{118.0} &  \multirow{2}{*}{230.4} &  \multirow{2}{*}{185.9} &  \multirow{2}{*}{132.1} &  \multirow{2}{*}{120.6} &  \multirow{2}{*}{155.1} &  \multirow{2}{*}{100.9} &  \multirow{2}{*}{113.0} &  \multirow{2}{*}{103.5} &  \multirow{2}{*}{131.0} &  \multirow{2}{*}{146.4}  \\ % hmn_209 (15)
& & & & & & & & & & & & & & & & & & \\  % Add this empty row to ensure multirow works correctly
\midrule
\multirow{1}{*}{MPL~\cite{MPL} + RCS} & 2 & 431.8 & 374.8 & 377.4 & 449.1 & 403.5 & 402.2 & 356.3 & 475.3 & 331.7 & 390.3 & 421.6 & 439.3 & 381.6 & 368.8 & 411.5 & 400.7 & 437.5 \\ % hmn_109 (14)
\multirow{1}{*}{MPL~\cite{MPL} + RCS + Rays} & 2 & 66.0 & 55.4 & 47.0 & 72.9 & 77.0 & 49.3 & 48.3 & 134.8 & 73.2 & 68.9 & 54.6 & 97.3 & 46.9 & 54.1 & 48.2 & 67.9 & 77.1 \\ % hmn_9 (13)
\midrule
\multirow{1}{*}{\textbf{RUMPL} (w/o Conf)} & 2 & 50.3 & 52.1 & 55.4 & 51.8 & 64.9 & 46.7 & 55.6 & 76.2 & 79.6 & 60.2 & 59.8 & 54.1 & 53.5 & 55.5 & 55.1 & 58.6 & 64.9 \\ % hrf_3126
\multirow{1}{*}{\textbf{RUMPL}} & 2 & \textbf{48.6} & \textbf{47.9} & \textbf{45.7} & \textbf{46.6} & \textbf{58.3} & \textbf{46.9} & \textbf{49.4} & \textbf{67.1} & \textbf{67.3} & \textbf{54.5} & \textbf{51.8} & \textbf{50.4} & \textbf{48.1} & \textbf{50.0} & \textbf{48.0} & \textbf{52.5} & \textbf{56.8} \\ % hrf_3125

\bottomrule
\end{tabular}
}

\end{table*}

\subsubsection{CMU Panoptic}
We follow \cite{voxelpose,wang2021mvp} and evaluate our RUMPL on the CMU dataset based on a subset of HD cameras (3, 6, 12, 13, 23). 
\cref{table:mpjpe-cmu} shows the MPJPE in $mm$ for different methods. In all methods, none of the images or 3D poses of the CMU dataset have been used in the training.

The results show that the off-the-shelf pre-trained methods all fail to generalize to the images in the CMU test set, even when enjoying higher number of views. 
\cref{table:mpjpe-cmu} confirms all the observations of \cref{table:mpjpe-h36m}.

\cref{table:mpjpe-cmu} confirms the observations of Table \cref{table:mpjpe-h36m} that the triangulation baseline does a good job when increasing the number of views to 5. However, using triangulation with only two views increases MPJPE by about $30\%$ compared to our MPL.

\begin{table}
\centering
\caption{3D MPJPE in $mm$ on CMU Panoptic test set between different methods. KP* indicates that the calculations were done only considering the keypoints most aligned with COCO as a fair comparison. The numbers in bold indicate lowest MPJPE with two views. RCS indicates that random camera setup is used to create the training dataset (using MHP). RUMPL uses RCS in all the cases. 'w/o Conf' indicates RUMPL without the 2D pose confidence.}
\label{table:mpjpe-cmu}
\resizebox{\columnwidth}{!}{
\begin{tabular}{l|c|c c}  
\toprule
\multirow{3}{*}{Method} & \multirow{3}{*}{\begin{tabular}[c]{@{}c@{}}\#\\ Views\end{tabular}} &  \multicolumn{2}{c}{$\downarrow$ MPJPE} \\ 
\cmidrule{3-4}
& &  (All KP) & (KP*) \\
\midrule
Learn.-triang. \cite{iskakov2019learnable} & 5 & - & 130.9 \\
PPT \cite{ma2022ppt} & 4 & 114.2 & 108.3 \\
\midrule
\multirow{1}{*}{Triangulation} & 2 & 43.0 & 44.0 \\
\midrule
% \midrule
\multirow{1}{*}{FullyConnected~\cite{kadkhodamohammadi2021generalizable} + RCS + Rays} & 2 & 108.6 & 127.4  \\ % crn_2009 (69)
\midrule
% \multirow{1}{*}{MPL} & 2 & 29.3 & 34.2  \\
\multirow{1}{*}{MPL~\cite{MPL} + RCS} & 2 & 274.9 & 278.8 \\ % crn_1909 (68)
\multirow{1}{*}{MPL~\cite{MPL} + RCS + Rays} & 2 & 45.4 & 55.7 \\ % crn_1809 (67)
\midrule
\multirow{1}{*}{\textbf{RUMPL} (w/o Conf)} & 2 & 36.6 & 41.1 \\ % crf_4926
\multirow{1}{*}{\textbf{RUMPL}} & 2 & \textbf{30.8} & \textbf{35.0} \\ % crf_4925

\bottomrule
\end{tabular}
}

\end{table}

\subsubsection{RICH}
% \textbf{RICH}:
\cref{table:mpjpe-rich} shows the 3D HPE results on the RICH val set. We evaluate our RUMPL on the RICH dataset by picking all the possible camera pairs of cameras in the scenes. 

When the displacement  of the person with respect to the room center (used as origin of the referential manipulated by RUMPL) gets large, it becomes helpful to translate the origin of the referential towards the human location before applying RUMPL. The inverse translation is then applied to the pose predicted by RUMPL. In practice, the translation vector is computed as a weighted average of the joints 3D coordinates obtained by triangulation. The weight associated to a keyppoint is defined as the minimum of the 2D confidence associated to this keypoint in the multiple views.
We name this process \textbf{Recentering}.

\cref{table:mpjpe-rich} shows how this technique improves the MPJPE.
It also confirms the observations from \cref{table:mpjpe-h36m} and \cref{table:mpjpe-cmu}. 

\begin{table}
\centering
\caption{3D MPJPE in $mm$ on RICH val set between different methods. KP* indicates that the calculations were done only considering the keypoints most aligned with COCO as a fair comparison. The numbers in bold indicate lowest MPJPE with two views. RCS indicates that random camera setup is used to create the training dataset (using MHP). RUMPL uses RCS in all the cases. 'w/o Conf' indicates RUMPL without the 2D pose confidence.}
\label{table:mpjpe-rich}
\resizebox{\columnwidth}{!}{
\begin{tabular}{l|c|c c}  
\toprule
\multirow{3}{*}{Method} & \multirow{3}{*}{\begin{tabular}[c]{@{}c@{}}\#\\ Views\end{tabular}} &  \multicolumn{2}{c}{$\downarrow$ MPJPE} \\ 
\cmidrule{3-4}
& &  (All KP) & (KP*) \\
\midrule
Learn.-triang. \cite{iskakov2019learnable} & 4 & - & 575.1 \\
\midrule
\multirow{1}{*}{Triangulation} & 2 & 59.7 & 54.5 \\
\midrule
% \midrule
\multirow{1}{*}{FullyConnected~\cite{kadkhodamohammadi2021generalizable} + RCS + Rays} & 2 & 191.0 & 211.4 \\ % rrn_209 (3)
\midrule
\multirow{1}{*}{MPL~\cite{MPL} + RCS} & 2 & 386.5 & 413.7 \\ % rrn_109 (2)
\multirow{1}{*}{MPL~\cite{MPL} + RCS + Rays} & 2 & 66.1 & 75.3 \\ % rrn_9 (1)
\midrule
\multirow{1}{*}{\textbf{RUMPL} (w/o Conf)} & 2 & 113.9 & 111.7 \\ % rrf_3426
\multirow{1}{*}{\textbf{RUMPL}} & 2 & 77.5 & 75.6 \\ % rrf_3425
\multirow{1}{*}{\textbf{RUMPL} + Recentering (w/o Conf)} & 2 & 57.9 & 56.7 \\ % rrf_3226
\multirow{1}{*}{\textbf{RUMPL} + Recentering} & 2 & \textbf{51.3} & \textbf{48.4} \\ % rrf_3225

\bottomrule
\end{tabular}
}

\end{table}

\subsection{Importance of 3D Ray Representation}
\label{ssec:importance-ray}

To show the importance of utilizing our proposed 3D ray representation, we consider three different scenarios where we change the input modality of the RUMPL.
The considered scenarios are when we feed the network with (i) 2D keypoints, (ii) 2D keypoints and camera calibration, and (iii) 3D ray.
\cref{table:importance-ray} shows that feeding the network with 3D rays improves the MPJPE (KP*) by 51.1\%, 59.7\%, and 88.1\% on Human3.6M, CMU, and RICH, respectively, compared to when 2D keypoints and camera calibration are used directly as input.

\begin{table}[h]
\centering
\caption{The importance of using our proposed 3D ray representation. Two views are considered during testing.}
\label{table:importance-ray}
\resizebox{\columnwidth}{!}{
\begin{tabular}{l|c c c}  
\toprule
\multirow{3}{*}{Input Modality of \textbf{RUMPL}} & \multicolumn{3}{c}{$\downarrow$ MPJPE (KP*) on} \\
& & & \\
& \cellcolor{blue!10}Human3.6M &  \cellcolor{blue!10}CMU & \cellcolor{blue!10}RICH \\
\midrule
\multirow{1}{*}{2D keypoints} & 695.7 & 269.0 & 428.8 \\
\multirow{1}{*}{2D keypoints + Camera Calibration} & 118.1 & 87.2 & 400.5 \\
\multirow{1}{*}{3D Ray} & \textbf{56.8} & \textbf{35.0} & \textbf{48.4} \\
\bottomrule
\end{tabular}
}

\end{table}

\subsection{Multi-person Extension}
Following MVGFormer \cite{liao2024multiple:mvgformer}, we test our RUMPL on multi-person CMU dataset by first matching the people in different views using a Hungarian method based on the average epipolar error of the keypoints.
It is worth noting that RUMPL needs to be retrained to account for occlusion scenarios occurring in a multi-person scenario.
We obtained AP100=74.2 compared to 35.7 and 64.9 for MVGFormer and triangulation, respectively.

\subsection{Comparison to Monocular Baselines}
\label{ssec:single-view baseline}

A straightforward approach to predict 3D pose from an arbitrary camera setup is to make use of an off-the-shelf monocular depth estimator coupled with the monocular 2D pose estimator.
In this section, we consider this method as a baseline to compare our RUMPL used in single-view fashion.

We have identified two state-of-the-art monocular depth estimators: Apple ml-depth-pro \cite{Bochkovskii2024:arxiv} and DepthAnything2 \cite{depth_anything_v1,depth_anything_v2}.
For this baseline, we first separately feed each image to the 2D pose estimator and the depth estimator. Then by combining the outputs, we get a 3D pose in world coordinates.
\cref{table:single-view} compares different methods on the three datasets when only one view is used.
It shows that even though RUMPL is not using depth as input, it still results in a better MPJPE compared to the baselines considered.
Extending the distance between the cameras and the human, as done when going from the CMU to the RICH dataset, obviously increases the estimation error.

\begin{table}[h]
\centering
\caption{\textbf{Monocular pose estimation}: comparison of 3D MPJPE (KP*) in $mm$ on different datasets.}
\label{table:single-view}
\resizebox{\columnwidth}{!}{
\begin{tabular}{l| c | c c c}  
\toprule
\multirow{3}{*}{Method} & \multirow{3}{*}{\# Views} & \multicolumn{3}{c}{$\downarrow$ MPJPE (KP*) on} \\
& & & \\
&& \cellcolor{blue!10}Human3.6M &  \cellcolor{blue!10}CMU & \cellcolor{blue!10}RICH \\
\midrule
\multirow{1}{*}{2D HPE + ml-depth-pro \cite{Bochkovskii2024:arxiv}} & 1 & 506.8 & 215.9 & 815.5 \\
\multirow{1}{*}{2D HPE + DepthAnything2 \cite{depth_anything_v2}} & 1 & 397.5 & 248.6 & 1135.5 \\
\midrule
\multirow{1}{*}{\textbf{RUMPL}} & 1 & \textbf{176.3} & \textbf{95.9} & \textbf{524.0} \\
\bottomrule
\end{tabular}
}

\end{table}

\subsection{Relevance of Our MHP}

To show the importance of utilizing our MHP to generate the RUMPL training set, we consider 3 different scenarios for training our RUMPL: 
A) training on the 3D and 2D-projected poses of Human3.6M (or CMU),
B) regressing the 3D poses from AMASS meshes, and training based on their corresponding projected 2D poses, and 
C) adopting our proposed MHP, \ie, regressing the 3D pose from AMASS and generating noisy 2D by running the 2D pose estimator on 2D mesh rendering.

\cref{table:importance-3dg} compares the 3D MPJPE of each scenario, when testing on Human3.6M, CMU, and RICH from left to right. We also include the baseline, \ie, triangulation, and only consider the keypoints in KP*, \ie, the keypoints for which the COCO’s format (same as the 2D pose estimator format in our study) is consistent with that of Human3.6M. The number of views for all scenarios is set to two, and the MPJPE is averaged on all possible pairs of cameras, selected among the ones that are commonly used in previous works, \ie, for Human3.6M: all four available cameras, for CMU: HD cameras (3, 6, 12, 13, 23), and for RICH: all available cameras.

By comparing the Scenario A and B, we observe a $53\%$ drop in MPJPE on Human3.6M dataset showing a benefit of using AMASS over CMU. 
However, when comparing the same scenarios on CMU and RICH dataset, we obsereve a slight increase in MPJPE showing that using AMASS over Human3.6M has no benefit by itself.
Interestingly, the use of our 2D-3D MHP pairs appears to significantly improve the MPJPE, up to $27.7\%$ compared to pairs derived from the datasets ground truths, thereby revealing that MPH is effective in enriching the training set.

\begin{table*}[t]
\centering
\caption{\textbf{The importance of MHP.} MPJPE values are reported in $mm$, the number of views is fixed to two. The testing dataset is Human3.6M, CMU, and RICH for the tables from left to right, respectively.}
\label{table:importance-3dg}
\begin{tabular}{ccc}
\begin{minipage}[t]{0.31\textwidth}
\vspace{0pt}
\centering

\resizebox{\textwidth}{!}{
\begin{tabular}{c | c | c c | c }
\multicolumn{5}{c}{\cellcolor{blue!10}On \textbf{Human3.6M}} \\
\toprule
\multirow{2}{*}{Method} & \multirow{2}{*}{Scenario}  & \multicolumn{1}{p{2cm}} {\centering Training \\ Dataset} & \multirow{2}{*}{MHP} & \multirow{2}{*}{\centering $\downarrow$ MPJPE}\\ 
\midrule
Triangulation & - & - & - & 114.7 \\
\midrule
\midrule
\multirow{3}{*}{\textbf{RUMPL} (ours)} 
& A  & CMU & \xmark & 167.1 \\ % w/o recentring
& B  & AMASS & \xmark & 78.6 \\
& C & AMASS & \cmark & \textbf{56.8} \\
\bottomrule
\end{tabular}
}
% \label{table:importance-3dg}
\end{minipage}
&
\begin{minipage}[t]{0.31\textwidth}
\vspace{0pt}
\centering
\resizebox{\textwidth}{!}{
\begin{tabular}{c | c | c c | c }
% \toprule
\multicolumn{5}{c}{\cellcolor{blue!10}On \textbf{CMU}} \\
\toprule
\multirow{2}{*}{Method} & \multirow{2}{*}{Scenario}  & \multicolumn{1}{p{2cm}} {\centering Training \\ Dataset} & \multirow{2}{*}{MHP} & \multirow{2}{*}{\centering $\downarrow$ MPJPE}\\ 
\midrule
Triangulation  & - & - & - & 44.0 \\
\midrule
\midrule
\multirow{3}{*}{\textbf{RUMPL} (ours)} 
& A & Human3.6M & \xmark & 43.1 \\ % chf_508
& B & AMASS & \xmark & 44.4 \\ % crf_4408
& C & AMASS & \cmark & \textbf{35.0} \\
\bottomrule
\end{tabular}
}
\end{minipage}
% \hspace{.2cm} % Adjust the width as needed
&
\begin{minipage}[t]{0.31\textwidth}
\vspace{0pt}
\centering
\resizebox{\textwidth}{!}{
\begin{tabular}{c | c | c c | c }
% \toprule
\multicolumn{5}{c}{\cellcolor{blue!10}On \textbf{RICH}} \\
\toprule
\multirow{2}{*}{Method} & \multirow{2}{*}{Scenario}  & \multicolumn{1}{p{2cm}} {\centering Training \\ Dataset} & \multirow{2}{*}{MHP} & \multirow{2}{*}{\centering $\downarrow$ MPJPE}\\ 
\midrule
Triangulation  & - & - & - & 54.5 \\
\midrule
\midrule
\multirow{3}{*}{\textbf{RUMPL} (ours)} 
& A & Human3.6M & \xmark & 56.9 \\ % rhf_408
& B & AMASS & \xmark & 57.6 \\ % rrf_3308
& C & AMASS & \cmark & \textbf{48.4} \\
\bottomrule
\end{tabular}
}

\end{minipage}

\end{tabular}

\end{table*}

\subsection{Inference speed}

To figure out whether our RUMPL can be a viable alternative to triangulation, it is worth considering its run time.
\cref{table:speed} shows the inference speed of RUMPL in frames/second where each frame is composed of all the input camera views of a single person. 
The numbers are measured on an NVidia A10 with a 32-core Intel Xeon Gold 6346 CPU running at 3.10GHz.
As shown in the table, RUMPL is able to run in real-time even when 5 views are used with the minimum batch size. Note that, while increasing the batch size improves the speed, it increases the delay to the output stream. 

\begin{table}[h]
\centering
\caption{The inference speed (frames/second) of RUMPL for different number of views and batch sizes.}

\label{table:speed}
% \scalebox{0.7}{
\resizebox{\columnwidth}{!}{
% \begin{tabular}{c|c c c c}
\begin{tabular}{>{\centering\arraybackslash}p{1cm} | >{\centering\arraybackslash}p{2cm} >{\centering\arraybackslash}p{2cm} >{\centering\arraybackslash}p{2cm} >{\centering\arraybackslash}p{2cm}}
\toprule
% \multicolumn{1}{p{1cm}|}{\centering \# \\ Views}
\multirow{2}{*}{\begin{tabular}{@{}c@{}}\# \\ Views\end{tabular}}
 & \multicolumn{4}{c}{Batch Size} \\
\cmidrule(lr){2-5}
  & 1 & 2 & 4 & 8 \\
\midrule
2 & 96.5 & 190.0 & 370.0 & 679.1 \\
3 & 95.1 & 190.0 & 368.9 & 699.2 \\
4 & 92.2 & 181.0 & 344.8 & 669.8 \\
5 & 89.3 & 162.4 & 292.0 & 467.6 \\
\bottomrule
\end{tabular}
}
% }

\end{table}

\subsection{Execution Time of Data Preparation and Training}

RUMPL needs to be trained with data prepared by MHP.
For the experiments in this paper, we make use of 128,109 meshes from AMASS. 
It approximately takes 21 hours to run MHP on 20 random cameras on 4 NVidia A10 with 32-core Intel Xeon Gold 6346 CPU running at 3.10GHz.

The RUMPL training time depends on the number of layers and heads used in the transformer network. 
For a network for which the maximum number of views is set to 5, with the configuration mentioned in the main text, it takes about 44 minutes to train RUMPL on data prepared by MHP. The training measurements are done on the same device using only 1 GPU.

\subsection{Qualitative Results}
\begin{figure*}[!t]
\centering
\resizebox{0.9\textwidth}{!}{
\subfloat[]{\includegraphics[trim={1cm 9cm 1cm 7cm},clip,width=\textwidth]{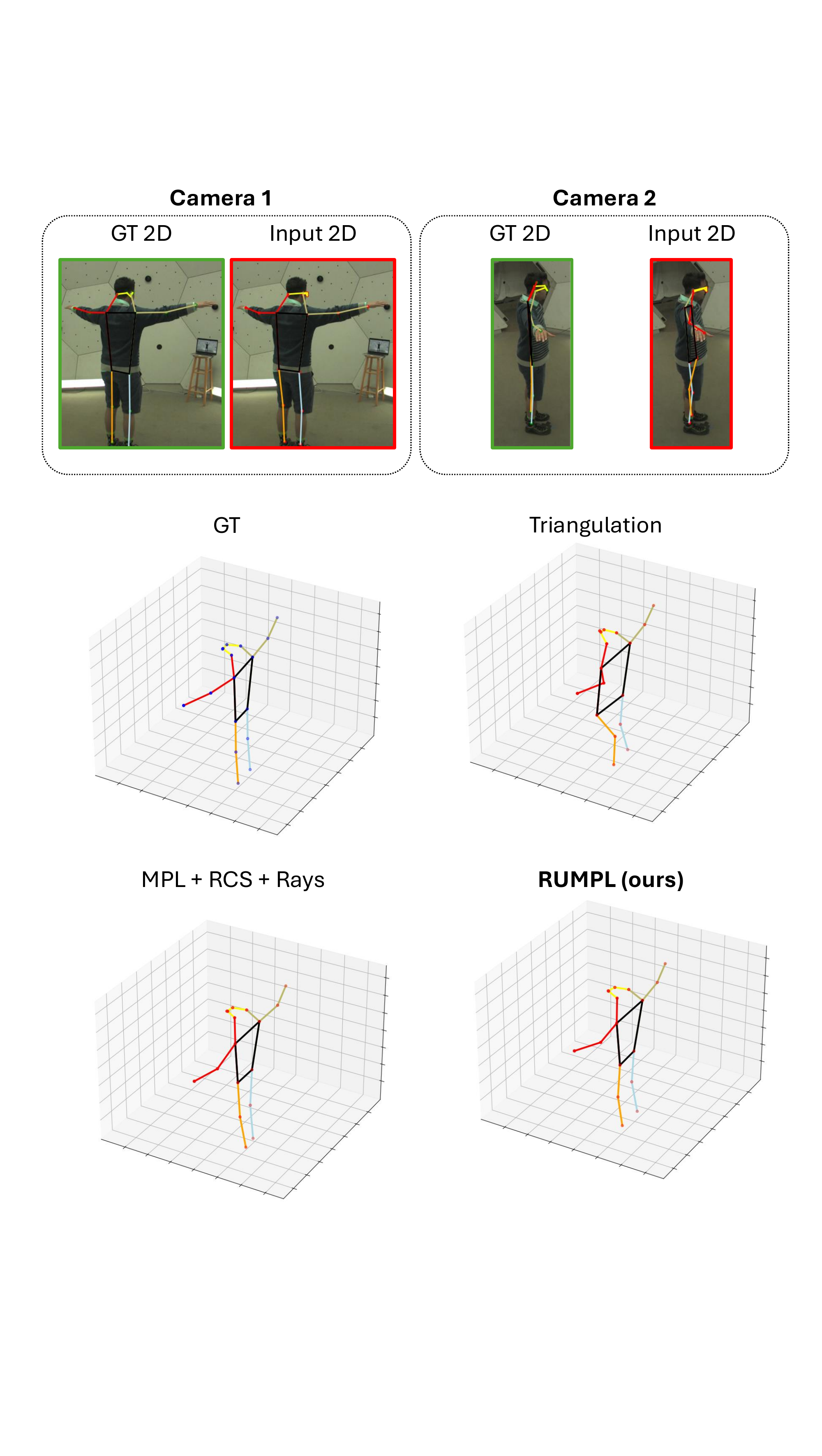}%
\label{fig_first_case}}
\hfil
\subfloat[]{\includegraphics[trim={1cm 9cm 1cm 7cm},clip,width=\textwidth]{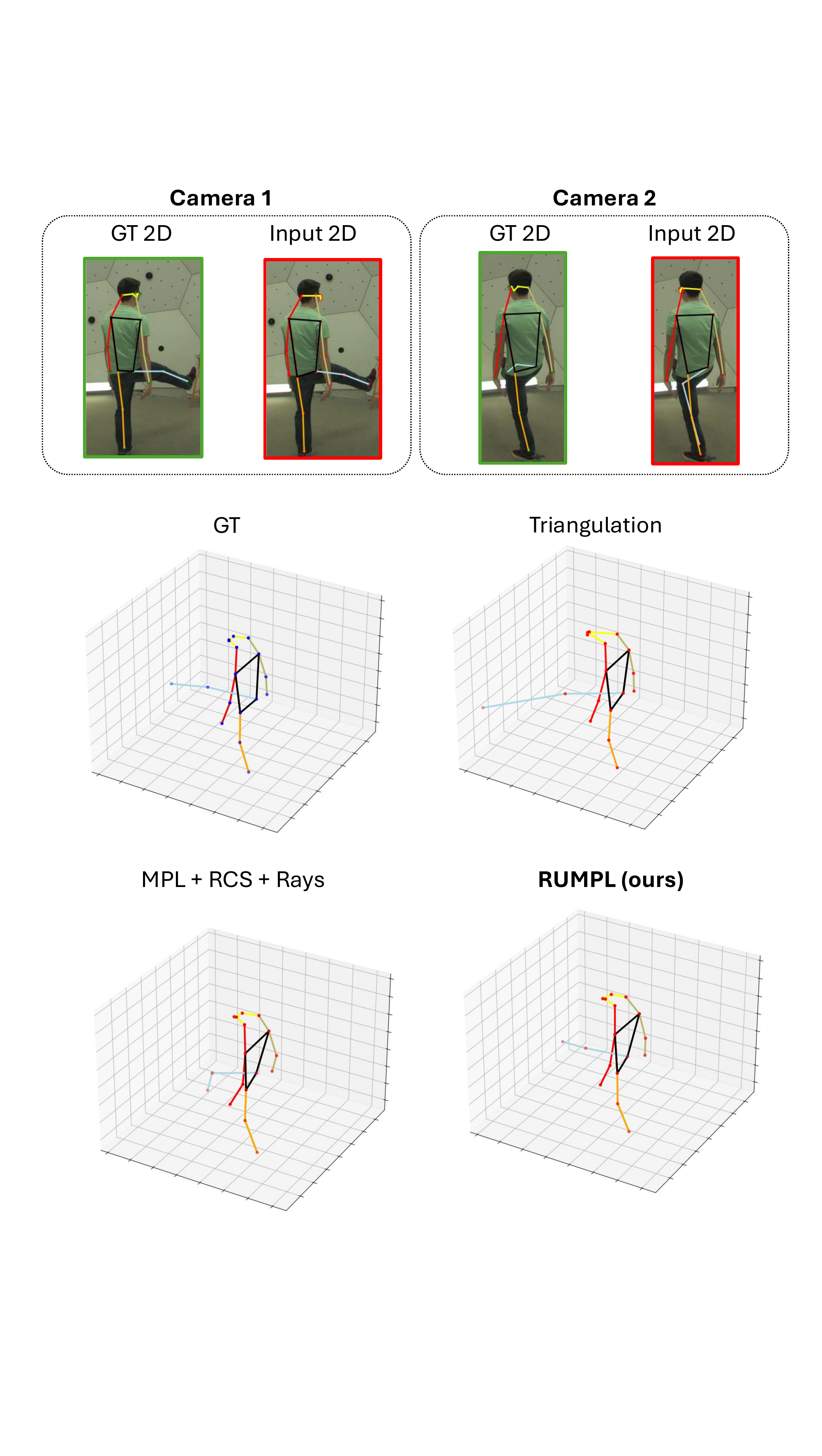}%
\label{fig_second_case}}
}
\caption{\textbf{Qualitative results.} In the top row, the GT and \textbf{MMPOSE} 2D represent the ground truth taken from the CMU dataset and the predicted 2D pose from the off-the-shelf 2D pose estimator, respectively. In the middle row, GT shows the 3D ground truth from the CMU dataset, and the other pose corresponds to the output of triangulation when the top row is given as inputs. In the bottom row, the outputs of MPL coupled with random camera setup (RCS) and ray representation and RUMPL are shown.}
\label{fig:qualitative}
\end{figure*}

\cref{fig:qualitative} shows some qualitative results comparing our RUMPL with triangulation and MPL \cite{MPL} coupled with random camera setup (RCS) and ray representation technique when methods are tested on CMU.

\subsection{Effect of Max Number of Views}

As mentioned in the main text, our RUMPL is compatible to any number of views as input thanks to its transformer view fusion scheme. However, during the training process, we have to set a maximum number of views. In other words, for each batch, we fix a number of views between 2 and a number associated to the maximum number of views. To study the effect of this parameter, we consider changing this parameter while testing it on 2 views from each dataset. \cref{table:max_number_views} shows the change in MPJPE (KP*) on the 3 different test sets for different values for maximum number of views used during training.

This table shows that this parameter has a neglectable impact on the accuracy of the RUMPL when the number of testing views is set to two.

\begin{table}[h]
\centering
\caption{Variation of MPJPE (KP*) when changing the maximum number of views used during training. Note that the test set used is always composed of two views.}
\label{table:max_number_views}

\resizebox{0.9\columnwidth}{!}{
\begin{tabular}{c|c c c}  
\toprule
\multirow{3}{*}{\begin{tabular}[c]{@{}c@{}}Max \# Views\\  (During Training)\end{tabular}} &  \multicolumn{3}{c}{$\downarrow$ MPJPE (KP*) on 2 Views} \\ 
\cmidrule{2-4}
& Human3.6M & CMU & RICH \\
\midrule
\multirow{1}{*}{2} & 76.0 & 35.0 & 46.7 \\
\multirow{1}{*}{3} & 61.0 & 34.7 & 46.7 \\
\multirow{1}{*}{4} & 62.2 & 34.2 & 48.1 \\
\multirow{1}{*}{5} & 56.8 & 35.0 & 48.4 \\
\multirow{1}{*}{6} & 61.3 & 35.2 & 49.2 \\
\multirow{1}{*}{7} & 59.7 & 34.2 & 49.7 \\
\multirow{1}{*}{8} & 57.4 & 35.9 & 50.0 \\
\multirow{1}{*}{9} & 57.4 & 36.1 & 51.1 \\
\multirow{1}{*}{10} & 57.7 & 37.8 & 50.9 \\
% \midrule

\bottomrule
\end{tabular}
}

\end{table}

\subsection{Effect of Camera Geometry on MPJPE}

To see the effect of the camera topoplogy, we design an experiment where we position the cameras on a sphere with different angles between them. As the range of camera angles in Human3.6M, CMU, and RICH datasets is very limited, we run this experiment on the test set of AMASS dataset (which is composed of CMU MoCap data). It goes without saying that we do not use this test set in any of our training scenarios. 
Here, we run MHP on both training and test set of AMASS. For the test set, we consider a spherical setup (of radius 6 $m$) with the center placed in $[0, 0, 0]$.
Then we place two cameras in a certain height with a specific angle. Next, we move one of them step by step and calculate the MPJPE on test set in the new camera setup. We repeat that for other heights. 
In other words, the two cameras are on a circle (at a given height on the sphere), and we change the angle between the cameras.
The heights are set to 2.2, 3.2, and 4 $m$. 

\begin{figure}[t]
    \centering
         \centering
         \includegraphics[scale=0.4]{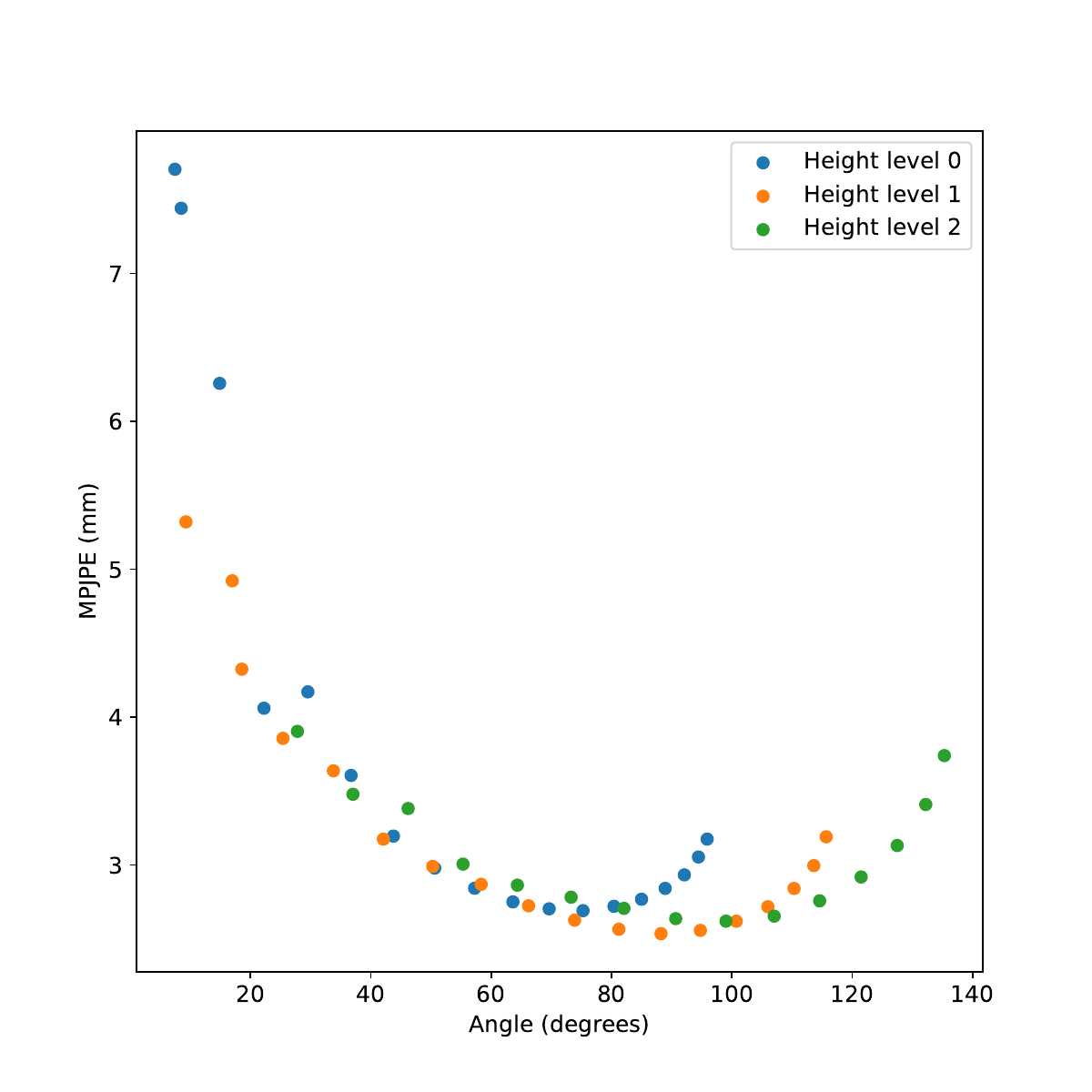}
    
    \caption{The change in MPJPE when changing the composition of the cameras. The angles are calculated in a 2-view camera setup. The cameras are placed on a sphere with radius of 6 $m$ and center of $[0,0,0]$. The heights are 2.2, 3.2, and 4 $m$.}
    % \vspace{-.3cm}
     \label{fig:mpjpe-angle}

\end{figure}

Then we train our RUMPL the training set and test the trained model on the test set (each time choosing the two views corresponding to the angle under experiment). \cref{fig:mpjpe-angle} shows the variation of MPJPE ($mm$) as a function of the angle ($\degree$) between the cameras.
This figure shows that when the angles get tighter or wider, the error goes higher.

\section{Conclusion}

We introduced a novel method for estimating 3D human poses from multiple calibrated views, addressing the major limitations of existing approaches; namely, their dependency on multi-view images paired with 3D ground truth, which are typically constrained to controlled laboratory settings.
By decoupling the process into two stages, \ie~robust 2D keypoint detection followed by 2D-to-3D pose lifting, we eliminate this dependency and enable generalization to arbitrary real-world conditions.
Our proposed transformer-based network, the Ray-based Universal Multi-view 3D Pose Lifter (RUMPL), represents 2D keypoints as 3D rays, allowing it to infer 3D poses directly in world coordinates. This formulation makes RUMPL independent of camera calibration parameters and the number of views, supporting universal deployment across different camera setups without retraining or fine-tuning. Furthermore, training RUMPL only requires synthetic 2D–3D pose pairs, avoiding the need for costly image–pose annotations.
Comprehensive experiments and ablation studies demonstrate that RUMPL achieves significant accuracy improvements over triangulation and state-of-the-art transformer-based methods, confirming its effectiveness, versatility, and scalability for practical, in-the-wild 3D human pose estimation.

\bibliographystyle{IEEEtran}
\bibliography{main}

% \bibitem{ref1}
% {\it{Mathematics Into Type}}. American Mathematical Society. [Online]. Available: https://www.ams.org/arc/styleguide/mit-2.pdf

% \bibitem{ref2}
% T. W. Chaundy, P. R. Barrett and C. Batey, {\it{The Printing of Mathematics}}. London, U.K., Oxford Univ. Press, 1954.

% \bibitem{ref3}
% F. Mittelbach and M. Goossens, {\it{The \LaTeX Companion}}, 2nd ed. Boston, MA, USA: Pearson, 2004.

% \bibitem{ref4}
% G. Gr\"atzer, {\it{More Math Into LaTeX}}, New York, NY, USA: Springer, 2007.

% \bibitem{ref5}M. Letourneau and J. W. Sharp, {\it{AMS-StyleGuide-online.pdf,}} American Mathematical Society, Providence, RI, USA, [Online]. Available: http://www.ams.org/arc/styleguide/index.html

% \bibitem{ref6}
% H. Sira-Ramirez, ``On the sliding mode control of nonlinear systems,'' \textit{Syst. Control Lett.}, vol. 19, pp. 303--312, 1992.

% \bibitem{ref7}
% A. Levant, ``Exact differentiation of signals with unbounded higher derivatives,''  in \textit{Proc. 45th IEEE Conf. Decis.
% Control}, San Diego, CA, USA, 2006, pp. 5585--5590. DOI: 10.1109/CDC.2006.377165.

% \bibitem{ref8}
% M. Fliess, C. Join, and H. Sira-Ramirez, ``Non-linear estimation is easy,'' \textit{Int. J. Model., Ident. Control}, vol. 4, no. 1, pp. 12--27, 2008.

% \bibitem{ref9}
% R. Ortega, A. Astolfi, G. Bastin, and H. Rodriguez, ``Stabilization of food-chain systems using a port-controlled Hamiltonian description,'' in \textit{Proc. Amer. Control Conf.}, Chicago, IL, USA,
% 2000, pp. 2245--2249.

% \end{thebibliography}

\newpage


\vfill

\end{document}